\pdfoutput=1

\documentclass[11pt]{article}

\usepackage[]{emnlp2021}

\usepackage{times}
\usepackage{latexsym}

\usepackage[T1]{fontenc}

\usepackage[utf8]{inputenc}

\usepackage{graphicx}
\usepackage{caption}
\usepackage{subcaption}
\usepackage{microtype}
\usepackage{listings}
\usepackage{xcolor}
\usepackage{hyperref}
\usepackage[hang,flushmargin]{footmisc}  
\usepackage[ruled,vlined]{algorithm2e}
\newcommand\ceil[1]{\lceil#1\rceil}
\newcommand\LN{\linebreak\noindent}

\title{ELIT: Emory Language and Information Toolkit}

\author{Han He \\
  Computer Science \\
  Emory University \\
  Atlanta GA 30322 USA \\
  \texttt{han.he@emory.edu} \\\And
  Liyan Xu \\
  Computer Science \\
  Emory University \\
  Atlanta GA 30322 USA \\
  \texttt{liyan.xu@emory.edu} \\\And
  Jinho D. Choi \\
  Computer Science \\
  Emory University \\
  Atlanta GA 30322 USA \\
  \texttt{jinho.choi@emory.edu}
}

\begin{document}
\maketitle

\begin{abstract}

We introduce ELIT, the Emory Language and Information Toolkit, which is a comprehensive NLP framework providing transformer-based end-to-end models for core tasks with a special focus on memory efficiency while maintaining state-of-the-art accuracy and speed.
Compared to existing toolkits, ELIT features an efficient Multi-Task Learning (MTL) model with many downstream tasks that include lemmatization, part-of-speech tagging, named entity recognition, dependency parsing, constituency parsing, semantic role labeling, and AMR parsing.
The backbone of ELIT's MTL framework is a\LN pre-trained transformer encoder that is shared across tasks to speed up their inference.
ELIT\LN provides pre-trained models developed on a remix of eight datasets. 
To scale up its service,\LN ELIT also integrates a RESTful Client/Server combination.
On the server side, ELIT extends its functionality to cover other tasks such as tokenization and coreference resolution, providing an end user with agile research experience.\LN
All resources including the source codes, documentation, and pre-trained models are publicly available at \url{https://github.com/emorynlp/elit}.


\end{abstract}
\section{Introduction}

The open source community has contributed many natural language processing (NLP) toolkits to the research and industry organizations, lowering the barrier of entry to access computational structures. 
Despite their wide usage, many NLP toolkits suffer from a major limitation that their architectures are bounded by the pipeline design \cite{manning2014stanford, straka2017tokenizing, gardner-etal-2018-allennlp, akbik2019flair, qi-etal-2020-stanza}, that leads to error propagation, large memory consumption, and high latency.\LN 
Although toolkits from the industry, such as spaCy, have started to exploit Multi-Task Learning (MTL), a lot of key components like semantic role labeling, constituency parsing and Abstract Meaning Representation parsing are not generally available. Additionally, they lack the ability to serve massive concurrent requests from the web, due to the lack of an efficient requests batching and corresponding multi-processing mechanism.

\begin{figure}[htbp!]
\centering
\includegraphics[width=\columnwidth]{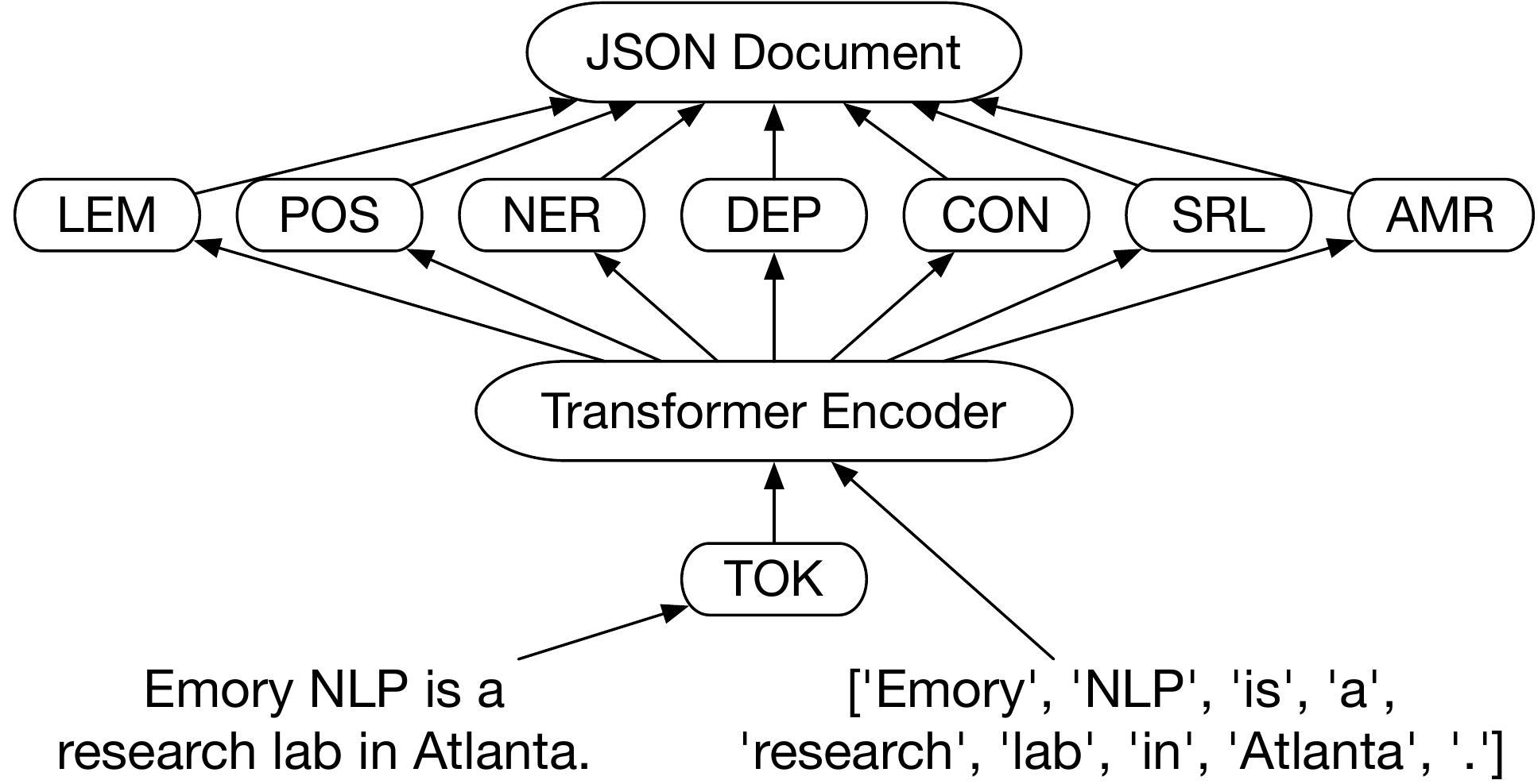}
\caption{The overview of the MTL framework in ELIT. ELIT takes input as either a plain text or a tokenized document, encodes it with a transformer encoder, and decodes multiple NLP tasks in parallel. The outputs of\LN NLP decoders are aggregated and presented as a JSON structure to the end user. Additionally, ELIT features a RESTful API for agile development.}

\label{fig:elit-architecture}
\end{figure}

\noindent In the face of these challenges, we introduce ELIT, an efficient yet accurate and fast NLP toolkit supporting the largest number of core NLP tasks with the boost of state-of-the-art transformer encoders.\LN
Compared to existing popular NLP toolkits, ELIT excels with the following outstanding advantages:

\begin{figure*}[htbp!]
\centering
\includegraphics[width=0.9\textwidth]{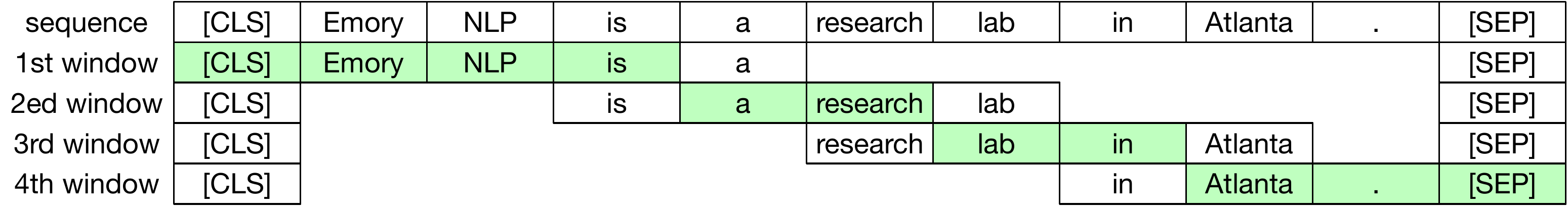}
\caption{A sliding window example for the sentence ``\textit{Emory NLP is a research lab in Atlanta .}'' with the maximum window size $m=6$. In each window, the tokens highlighted in green will be used in the ultimate outputs.}
\label{fig:sw}
\end{figure*}

\begin{itemize}
  \item \textbf{MTL Framework} ELIT is powered by an efficient MTL framework to accommodate many NLP tasks, spanning from surface tasks, syntactic tasks to semantic tasks.
  
  \item \textbf{State-of-the-Art Performance} Backed up by the latest transformer encoder and decoders, ELIT establishes state-of-the-art on most tasks and yields comparable results on the others.
  
  \item \textbf{Requests Batching for Concurrency} In order to scale up its inference, ELIT provides a built-in multi-worker web server featured with an efficient requests batching mechanism. 
\end{itemize}

\noindent ELIT embraces the NLP community with a fully open source license and lots of pre-trained models for public download. We hope ELIT can facilitate NLP research and applications, and bring the benefits of the NLP techniques to broader audiences.

\section{System Architecture}

From the user's view, ELIT provides 2 sets of APIs: (1) a native multi-learning task framework coupled with a rule-based tokenizer (see Figure~\ref{fig:elit-architecture}); (2) a RESTful client interface to the MTL server. 
In this section, we introduce their architecture designs.

\subsection{Architecutre}

\paragraph{Multi-Task Learning (MTL)} Given a tokenized sentence (or document), ELIT employs a sub-word tokenizer to further tokenize each token into sub-tokens. These sub-tokens are fed into a fine-tuned Transformer Encoder (TE) to get their contextualized embeddings. For those sentences longer than the maximum input length of the TE, a sliding window routine is invoked to mitigate the issue. 

\begin{algorithm}
  \DontPrintSemicolon
  \SetKwFunction{sw}{sliding\_window}
  \SetKw{KwBy}{by}
  \SetKwProg{Pn}{Function}{:}{\KwRet}
  \Pn{\sw{$\mathbf{t}_{n \times 1}, m$}}{
  		$s \gets \ceil{\frac{m - 2}{2}}$ \tcp*{stride}
  		$\mathbf{w} \gets$ \texttt{[]} \tcp*{windows}
  		\For{$i\gets 1$ \KwTo $n-1$ \KwBy $s$}{
    		$\mathbf{w} \gets \mathbf{w} \oplus \mathbf{t}_{0:1} \oplus \mathbf{t}_{i:i+s} \oplus \mathbf{t}_{n-1:n} $
	    }
        \KwRet\ $\mathbf{w}$
  } 
  \SetKwFunction{rs}{restore} 
   \Pn{\rs{$\mathbf{w}_{l \times 1}, m$}}{
  		$s \gets \ceil{\frac{m - 2}{2}}$ \tcp*{stride}
  		$o \gets \ceil{\frac{s}{2}}$  \tcp*{start offset}
  		$\mathbf{r} \gets$ \texttt{[]} \tcp*{restored sequence}
  		\For{$i\gets 0$ \KwTo $l$ \KwBy $1$}{
  		    \If{$i < o \lor o \le i \bmod m < o+s \lor \lfloor \frac{l}{m} \rfloor \times m+o \le i$}{ 
         		$\mathbf{r} \gets \mathbf{r} \oplus \mathbf{w}_{i:i+1}$
		    } 
	    }
        \KwRet\ $\mathbf{r}$
  } 
 \caption{Sliding Window}
 \label{alg:sw}
\end{algorithm}

\noindent As in Algorithm~\ref{alg:sw}, the \texttt{sliding\_window} sub-routine takes input as a list of sub-tokens $\mathbf{t}$ and slices them into windows with the maximal size $m$, which will be fed into a Transformer Encoder. 
Then the hidden states are restored using the \texttt{restore} sub-routine to match the original sequence such that in each window the inner parts will be used (Figure~\ref{fig:sw}).
Then, average pooling is applied to sub-token embeddings to get the corresponding token embeddings. 
Finally, the token embeddings are fed into the decoder of every task in parallel. 
This MTL architecture apportions the cost of TE between decoders, yielding lower overall latency than running TE individually for each of the decoders. 
It also reduces the training cost and deployment efforts due to its compact structure.

\begin{figure}[htbp!]
\centering
\includegraphics[width=\columnwidth]{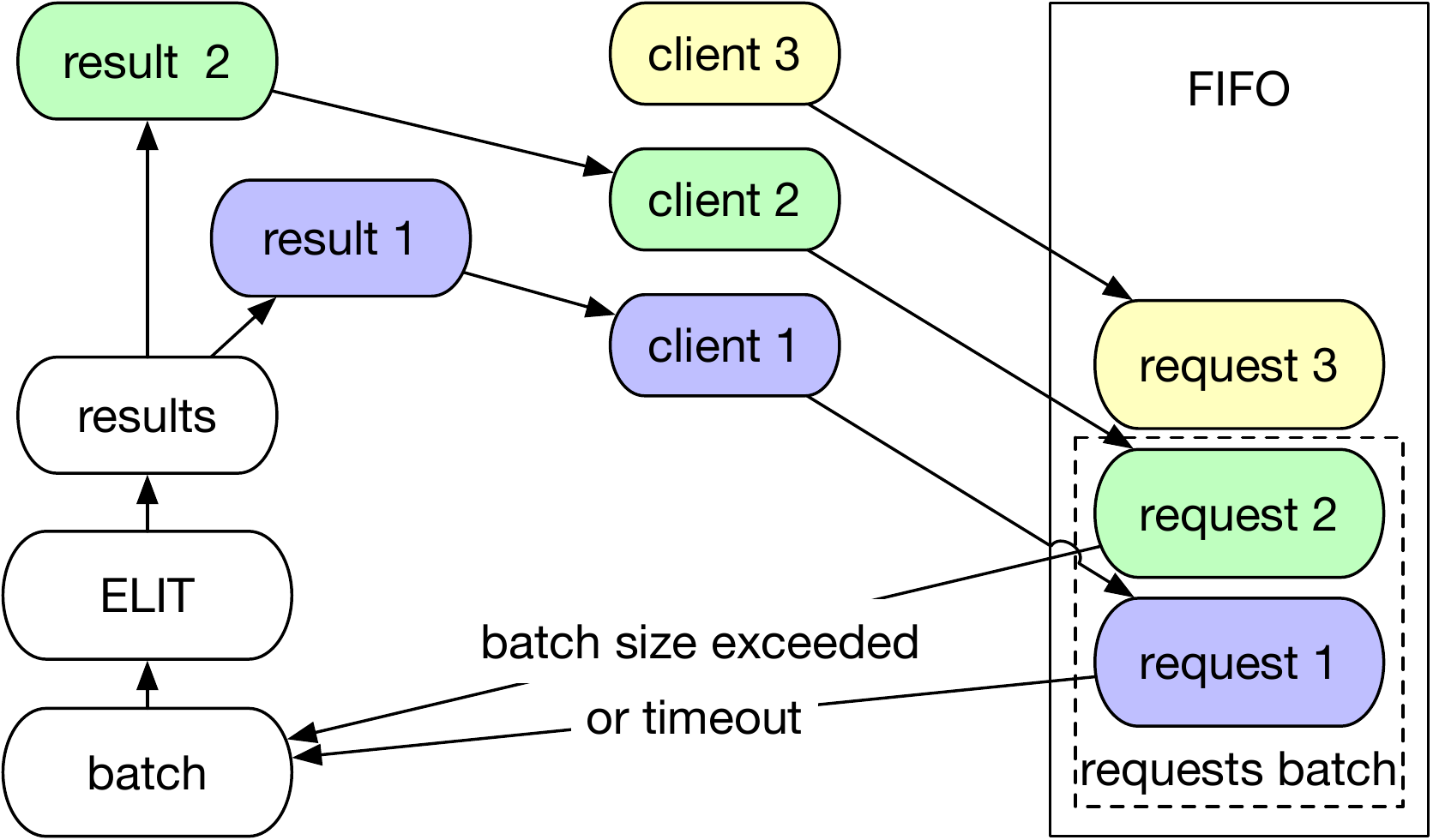}
\caption{System diagram of the ELIT RESTful server.}
\label{fig:elit-server}
\end{figure}

\paragraph{RESTful API} In scenarios with strict latency requirements such as a dialogue system or real-time machine translator, not even the native MTL can respond timely to highly concurrent requests due to the Global Interpreter Lock (GIL) of Python\footnote{We also investigated serving techniques free of GIL such as PyTorch JIT and ONNX. However, these techniques usually convert the model into a symbolic (static) representation, hampering features such as dynamic tasks scheduling and length-based batching. Thus, we leave them for future work.}. 
To scale up the inference, we also implement a RESTful Client/Server with requests batching in ELIT as illustrated by Figure~\ref{fig:elit-server}.

\noindent On the server side, HTTP requests are bucketed into batches according to their arrival time, and predicted concurrently by several worker processes spawned over the CPUs and GPUs. 
It is possible that sentences from different requests are put into the same batch, as long as the requested tasks are the same. When no worker is idle, a First-In-First-Out (FIFO) queue is used to store the new coming requests. On the client side, the requests batching mechanism is transparent and each user is able to parse their documents in the same way as using a native MTL API exclusively. Note that both APIs share similar semantics and return the same JSON format so that users can easily switch APIs without breaking changes.

\subsection{Decoders}

Within the MTL architecture, ELIT provides a set of decoders for widely used NLP tasks. Each task is supported by a state-of-the-art decoder. In this section, we will briefly introduce each of them.

\paragraph{LEM} ELIT reduces the lemmatization problem to a sequence tagging problem by predicting a tag for each token representing an edit script to transform the token form to its lemma \citep{chrupala2006simple, muller2015joint, kondratyuk-straka-2019-75}.

\paragraph{POS} For efficiency, a linear layer is used to predict the part-of-speech tags. We acknowledge the effectiveness of character level and case features \citep{bohnet-etal-2018-morphosyntactic,akbik-etal-2018-contextual} for POS, however, the improvements to accuracy brought by these features are marginal compared to the latency they introduce in our setting. Thus, we do not integrate them.

\paragraph{NER} A biaffine layer \citep{dozat:17a} is used for the NER task. Different from \cite{yu-etal-2020-named}, we avoid using document level features or the variational BiLSTM for faster decoding speed.

\paragraph{DEP} Two biaffine layers \citep{dozat:17a} are used to compute the dependent-head score matrix and label distribution. Then the Chu-Liu/Edmonds algorithm \cite{chu1965shortest, edmonds1967optimum} is applied on the score matrix to decode the maximum spanning tree.

\paragraph{CON} The two-stage CRF \citep{ijcai2020-560} decoder is used for CON which is optimized using a tree-structure CRF objective on unlabeled constituents. POS features are avoided to remove the dependency on the POS decoder.

\paragraph{SRL} The end-to-end span ranking decoder \citep{he-etal-2018-jointly} is used for SRL. Their attention-based span representations are replaced by average pooled embeddings for simplicity. 

\paragraph{AMR} The graph sequence transduction decoder \citep{he-choi:2021:iwpt} is used for AMR. Linguistic features including POS, NER and LEM embeddings are removed for efficiency. 

\noindent Separated from the MTL architecture, ELIT also provides two coreference resolution components that perform the traditional document-level decoding (DCR) as well as online decoding (OCR).

\paragraph{DCR} We adapt the implementation from \citet{xu-choi-2020-revealing} that is based on the end-to-end coreference resolution system with transformer encoders \citep{joshi-etal-2019-bert,joshi-etal-2020-spanbert}.

\paragraph{OCR} The model takes a current utterance and conversation context as the input, extracting mentions in the current utterance (including singletons), and resolving their coreference with previously predicted mentions from the past history.

\section{System Usage}

\subsection{MTL API}

The user interface of ELIT is designed to hide the underlying complexity from end users, allowing them for agile developments of NLP models.
To fulfill this goal, ELIT packs all sophisticated work required by other NLP toolkits, such as downloading models, using the right class to create an instance, loading and deploying it to GPU where possible, feeding the outputs from one component to another, into the following 3 lines of codes:

\begin{lstlisting}
import elit
nlp = elit.load('LEM_POS_NER_DEP_SDP_CON_AMR_EN')
doc = nlp(['Emory','NLP','is','in','Atlanta'])
\end{lstlisting}

\noindent As indicated by the identifier passed to the \texttt{load} call, \texttt{LEM\_POS\_NER\_DEP\_SDP\_CON\_AMR\_EN} takes input sentences and performs LEM, POS, NER, DEP, SDP, CON and AMR jointly.
The interface for coreference resolution is the same as MTL except that the order of input sentences is expected to be consistent with the document.
The following snippet loads the coreference model with SpanBERT\textsubscript{Large} and perform predictions for DCR:

\begin{lstlisting}
import elit
nlp = elit.load('DOC_COREF_SPANBERT_LARGE_EN')
doc = nlp([['Emory','NLP','is','in','Atlanta'],
           ['It','is','founded','in','2014']])
\end{lstlisting}

\subsection{RESTful API}

In a more common setting where multiple users require their documents to be parsed by ELIT, its web server can be set up to handle concurrent requests efficiently:

\begin{lstlisting}
!elit serve
import elit
nlp = Client('http://0.0.0.0:8000')
doc = nlp(['Emory NLP is in Atlanta'])
\end{lstlisting}

\noindent Note that we design the native MTL API and the RESTful API such that they share the same semantics and can be used interchangeably.
The only subtle difference is that, the native MTL accepts only tokenized sentences whereas the RESTful API additionally accepts raw text which will be tokenized, segmented to sentences on the server side.

\subsection{Output Format}

In both MTL and RESTful APIs, a \texttt{Document} instance will be returned to the user, which is a Python \texttt{dict} storing all annotation results. For each task, its annotations are associated to a key indicated by its task name, e.g., the above \texttt{doc} will have the following structure:
\begin{lstlisting}
{
  "tok": [
    ["Emory", "NLP", "is", "in", "Atlanta"]
  ],
"lem": [
    ["emory", "nlp", "be", "in", "atlanta"]
  ],
  "pos": [
    ["NNP", "NNP", "VBZ", "IN", "NNP"]
  ],
  "ner": [
    [["ORG", 0, 2, "Emory NLP"], ["GPE", 4, 5, "Atlanta"]]
  ],
  "srl": [
    [[["ARG1", 0, 2, "Emory NLP"], ["PRED", 2, 3, "is"], ["ARG2", 3, 5, "in Atlanta"]]]
  ],
  "dep": [
    [[1, "com"], [3, "nsbj"], [3, "cop"], [-1, "root"], [3, "obj"]]
  ],
  "con": [
    ["TOP", [["S", [["NP", [["NNP", ["Emory"]], ["NNP", ["NLP"]]]], ["VP", [["VBZ", ["is"]], ["PP", [["IN", ["in"]], ["NP", [["NNP", ["Atlanta"]]]]]]]]]]]]
  ],
  "amr": [
    [["c0", "ARG1", "c1"], ["c0", "ARG2", "c2"], ["c0", "instance", "be-located-at-91"], ["c1", "instance", "emory nlp"], ["c2", "instance", "atlanta"]]
  ],
}
\end{lstlisting}

\begin{itemize}
	\item \texttt{tok} stores the surface form of each token.
	\item \texttt{lem} stores the lemmatization of each token.
	\item \texttt{pos} stores the part-of-speech tag of each token. In this component, the Penn Treebank Part-of-speech tags \cite{santorini1990part} are used.
	\item \texttt{ner} stores the \texttt{(type, start, end,  form)} of each entity. In this component, the OntoNotes 5 NER annotations \cite{weischedel2013ontonotes} are used.
	\item \texttt{srl} stores the \texttt{(role, start, end, form)} of the predicates and arguments corresponding to each flattened predicate-argument structure. In this component, the OntoNotes 5 SRL annotations \cite{weischedel2013ontonotes} are used with an additional role \texttt{PRED} indicating the predicate.
	\item \texttt{dep} stores the \texttt{(head, relation)} of each token, with the offset starting from -1 (ROOT). In this component, the primary dependency of Deep Dependency Graph Representation \cite{choi2017deep} is used. The full representation with secondary dependencies are provided in another component.
	\item \texttt{con} stores the constituency trees, specifically \texttt{(label, child-constituents)} for the non-terminal constituents and \texttt{form} for the terminals. Note that we designed a nested list representation with the round brackets replaced by square brackets to avoid ambiguity and make it compatible with JSON. When not being printed out, the \texttt{Document} class will convert our nested list structure to the conventional bracketed tree.
	\item \texttt{amr} stores the logical triples of Abstract Meaning Representation in the format of \texttt{(source, relation, target)}. Note that the \texttt{Document} class will convert it to Penman format \cite{goodman2020penman} when being accessed through code.
\end{itemize}

\noindent For the DCR example shown before, the following output should be generated:

\begin{lstlisting}
{
  "dcr": [
    [[0, 0, 2, "Emory NLP"], [1, 0, 1, "It"]]]
  ]
}
\end{lstlisting}
where \texttt{dcr} contains a list of clusters, and each cluster consists of spans referring to the same entity, in the format of \texttt{(sentence-index, token-start, token-end, text)}. The input and output formats for OCR are shown in details on the GitHub page \footnote{\url{https://github.com/emorynlp/elit/blob/main/docs/data_format_ocr.md}}.

\subsection{Training}



In ELIT, the APIs to train a new model are as strait-forward as the inference APIs. Instead of resorting to a configuration file, we opt for the native Python APIs which offer built-in documentations of each parameter and the necessary type checking. Sticking to the Python APIs requires no extra efforts to learn another language to write config files. These benefits are illustrated in the following code snippet, which demonstrates how to train a joint NER and DEP component with easy integration with the RoBERTa encoder:

\begin{lstlisting}
tasks = {
    'ner': BiaffineNamedEntityRecognition(
        'srl/train.english.v4.jsonlines',
        'srl/development.english.v4.jsonlines',
        'srl/test.english.v4.jsonlines',
        SortingSamplerBuilder(batch_size=128, batch_max_tokens=12800),
        lr=5e-3,
    ),
    'dep': BiaffineDependencyParsing(
        'ddr/trn.conllu',
        'ddr/dev.conllu',
        'ddr/tst.conllu',
        SortingSamplerBuilder(batch_size=128, batch_max_tokens=12800),
        lr=1e-3,
    ),
}
mtl = MultiTaskLearning()
mtl.fit(
    ContextualWordEmbedding(
        'token',
        "roberta-base",
        average_subwords=True,
        max_sequence_length=512,
        word_dropout=[.2, 'unk'],
    ),
    tasks,
    save_dir,
    epochs=30,
    encoder_lr=5e-5,
)
mtl.evaluate(save_dir)	
\end{lstlisting}

\noindent For each task, its popular training file format (e.g., \texttt{tsv}, \texttt{CONLL-U}) is well retained such that users can make use of the readily available open access datasets on the web. Note that our \texttt{SortingSamplerBuilder} builds a sampler which groups sequences similar in length into the same batch such that training and inference will be significantly accelerated. The training for coreference resolution models are performed outside ELIT, for which users can refer to our documentation.

\section{Performance Evaluation}

\subsection{Datasets and Metrics}

\paragraph{Datasets} The ELIT models are trained on a mixture of OntoNotes 5 \citep{weischedel2013ontonotes}, BOLT English Treebanks \citep{song2019bolt, tracey2019bolt}, THYME/SHARP/MiPACQ Treebanks \citep{albright2013towards}, English Web Treebank \citep{bies2012english}, Questionbank \citep{judge2006questionbank} and AMR 3.0 dataset \citep{knight2020abstract}. Batches of NLP tasks are mixed together so that even if a corpus offers no annotation for some tasks, it can still be exploited by ELIT. For LEM, POS, NER, DEP, CON and SRL, the mixed corpora are split into training, development and test set with a 8:1:1 ratio. For AMR, the standard splits are used.

\paragraph{Metrics} The following evaluation are used for each task - LEM and POS: accuracy, NER: span-level labeled F1, DEP: labeled attachment score, CON: constituency-level labeled F1, SRL: micro-averaged F1 of predicate-argument-label triples, AMR: Smatch \citep{cai2013smatch}, DCR/OCR: Averaged F1 of MUC, B\textsuperscript{3}, and CEAF\textsubscript{$\phi_4$}.

\begin{table}[h]
    \begin{subtable}[h]{\columnwidth}
        \centering
        \begin{tabular}{l | l }
        Model & F1  \\
        \hline \hline
        Stanza \cite{qi-etal-2020-stanza} & 88.8  \\
        Flair \cite{akbik2019flair} & 89.7 \\
        spaCy RoBERTa  & 89.8  \\
        biaffine w/o doc \cite{yu-etal-2020-named} & 89.8$^\dagger$ \\
        BART-NER \cite{yan2021unified} & 90.4 \\
        \hline
        ELIT BERT-large & 89.7 \\
       \end{tabular}
       \caption{NER performance on OntoNotes 5. Scores with $^\dagger$ are from experiments by \citet{yan2021unified} excluding document context.}
       \label{tab:ner}
    \end{subtable}
    \vspace{1em}

    \begin{subtable}[h]{\columnwidth}
        \centering
        \begin{tabular}{l | l }
         Model & F1  \\
        \hline \hline
        e2e \cite{li2019dependency} & 83.1  \\   
        SpanRel \cite{jiang-etal-2020-generalizing} & 82.4  \\   
        \hline
        ELIT BERT-large & 84.0 \\
       \end{tabular}
       \caption{Span-based end-to-end SRL results on CoNLL'12.}
        \label{tab:srl}
     \end{subtable}
     \vspace{1em}

    \begin{subtable}[h]{\columnwidth}
        \centering
        \begin{tabular}{l | l }
         Model & F1  \\
        \hline \hline
        SpanBERT \citep{joshi-etal-2020-spanbert} & 79.6  \\   
        Higher-Order \cite{xu-choi-2020-revealing} & 80.2  \\   
        \hline
        ELIT SpanBERT-large & 80.2 \\
       \end{tabular}
       \caption{Coreference resolution (DCR) results on CoNLL'12.}
        \label{tab:coref}
     \end{subtable}
     \caption{Performance on OntoNotes 5, CoNLL-2012.}
     \label{tab:ontonotes}
\end{table}

\subsection{Training}

\begin{table*}[htbp!]
\centering
\begin{tabular}{c||c|c|c|c|c|c|c||c}
            & LEM   & POS   & NER   & DEP   & CON   & SRL   & AMR  & Samples/Sec. \\ \hline \hline
MTL-RoBERTa & 99.65 & 98.08 & 89.01 & 91.21 & 90.78 & 77.78 & 73.9 & 264   \\ 
\end{tabular}
\caption{Performance and speed of the MTL-RoBERTa in ELIT.}
\label{tab:mtl-performance}
\end{table*}

All hyper-parameters are tuned on the development set of OntoNotes 5 and applied to the mixed datasets. For the pre-trained transformer encoder, Electra base \cite{clark2020electra} and RoBERTa base \cite{liu2019roberta} are compared and we opt for RoBERTa due to its better results on more tasks.

\subsection{Accuracies and Speed}

The ELIT MTL model is trained and evaluated using a single TITAN RTX GPU for 30 epochs, which take roughly 36 hours. The scores on test set and decoding speed are listed in Table~\ref{tab:mtl-performance}.

To make relatively fair comparisons with the existing toolkits and models from published work, we train single-task learning models on OntoNotes 5 and CoNLL-2012 as well with the standard split for NER, SRL and Coref. The results are shown in Table \ref{tab:ontonotes}. We find that ELIT achieved either higher or close performance in comparison to existing toolkits or models.

\section{Conclusion and Future Work}

We introduced ELIT, the Emory Language and Information Toolkit providing the largest number of NLP tasks within an efficient MTL framework. Our APIs demonstrate the usability of ELIT within several function calls. The interchangeable RESTful API further extends ELIT with extra features and enables it to serve large-scale concurrent requests. 
For future work, we plan the followings:

\begin{itemize}
	\item Train multilingual models since the design of ELIT is language agonistic. 
	\item Integrate statistical tokenizers to replace the rule-based one and to support non-tokenized languages like Korean and Chinese.
	\item Exploit model distillation and compression techniques to reduce the size of transformer encoders.
\end{itemize}


\bibliography{emnlp2021}

\begin{thebibliography}{36}
\expandafter\ifx\csname natexlab\endcsname\relax\def\natexlab#1{#1}\fi

\bibitem[{Akbik et~al.(2019)Akbik, Bergmann, Blythe, Rasul, Schweter, and
  Vollgraf}]{akbik2019flair}
Alan Akbik, Tanja Bergmann, Duncan Blythe, Kashif Rasul, Stefan Schweter, and
  Roland Vollgraf. 2019.
\newblock Flair: An easy-to-use framework for state-of-the-art nlp.
\newblock In \emph{Proceedings of the 2019 Conference of the North American
  Chapter of the Association for Computational Linguistics (Demonstrations)},
  pages 54--59.

\bibitem[{Akbik et~al.(2018)Akbik, Blythe, and
  Vollgraf}]{akbik-etal-2018-contextual}
Alan Akbik, Duncan Blythe, and Roland Vollgraf. 2018.
\newblock \href {https://www.aclweb.org/anthology/C18-1139} {Contextual string
  embeddings for sequence labeling}.
\newblock In \emph{Proceedings of the 27th International Conference on
  Computational Linguistics}, pages 1638--1649, Santa Fe, New Mexico, USA.
  Association for Computational Linguistics.

\bibitem[{Albright et~al.(2013)Albright, Lanfranchi, Fredriksen, Styler,
  Warner, Hwang, Choi, Dligach, Nielsen, Martin et~al.}]{albright2013towards}
Daniel Albright, Arrick Lanfranchi, Anwen Fredriksen, William~F Styler, Colin
  Warner, Jena~D Hwang, Jinho~D Choi, Dmitriy Dligach, Rodney~D Nielsen, James
  Martin, et~al. 2013.
\newblock Towards comprehensive syntactic and semantic annotations of the
  clinical narrative.
\newblock \emph{Journal of the American Medical Informatics Association},
  20(5):922--930.

\bibitem[{Bies et~al.(2012)Bies, Mott, Warner, and Kulick}]{bies2012english}
Ann Bies, Justin Mott, Colin Warner, and Seth Kulick. 2012.
\newblock English web treebank.
\newblock \emph{Linguistic Data Consortium, Philadelphia, PA}.

\bibitem[{Bohnet et~al.(2018)Bohnet, McDonald, Sim{\~o}es, Andor, Pitler, and
  Maynez}]{bohnet-etal-2018-morphosyntactic}
Bernd Bohnet, Ryan McDonald, Gon{\c{c}}alo Sim{\~o}es, Daniel Andor, Emily
  Pitler, and Joshua Maynez. 2018.
\newblock \href {https://doi.org/10.18653/v1/P18-1246} {Morphosyntactic tagging
  with a meta-{B}i{LSTM} model over context sensitive token encodings}.
\newblock In \emph{Proceedings of the 56th Annual Meeting of the Association
  for Computational Linguistics (Volume 1: Long Papers)}, pages 2642--2652,
  Melbourne, Australia. Association for Computational Linguistics.

\bibitem[{Cai and Knight(2013)}]{cai2013smatch}
Shu Cai and Kevin Knight. 2013.
\newblock Smatch: an evaluation metric for semantic feature structures.
\newblock In \emph{Proceedings of the 51st Annual Meeting of the Association
  for Computational Linguistics (Volume 2: Short Papers)}, pages 748--752.

\bibitem[{Choi(2017)}]{choi2017deep}
Jinho Choi. 2017.
\newblock Deep dependency graph conversion in english.
\newblock In \emph{TLT}, pages 35--62.

\bibitem[{Chrupa{\l}a(2006)}]{chrupala2006simple}
Grzegorz Chrupa{\l}a. 2006.
\newblock Simple data-driven context-sensitive lemmatization.

\bibitem[{Chu(1965)}]{chu1965shortest}
Yoeng-Jin Chu. 1965.
\newblock On the shortest arborescence of a directed graph.
\newblock \emph{Scientia Sinica}, 14:1396--1400.

\bibitem[{Clark et~al.(2020)Clark, Luong, Le, and Manning}]{clark2020electra}
Kevin Clark, Minh-Thang Luong, Quoc~V Le, and Christopher~D Manning. 2020.
\newblock Electra: Pre-training text encoders as discriminators rather than
  generators.
\newblock \emph{arXiv preprint arXiv:2003.10555}.

\bibitem[{Dozat and Manning(2017)}]{dozat:17a}
Timothy Dozat and Christopher~D. Manning. 2017.
\newblock \href {https://openreview.net/pdf?id=Hk95PK9le} {{Deep Biaffine
  Attention for Neural Dependency Parsing}}.
\newblock In \emph{Proceedings of the 5th International Conference on Learning
  Representations}, ICLR'17.

\bibitem[{Edmonds(1967)}]{edmonds1967optimum}
Jack Edmonds. 1967.
\newblock Optimum branchings.
\newblock \emph{Journal of Research of the national Bureau of Standards B},
  71(4):233--240.

\bibitem[{Gardner et~al.(2018)Gardner, Grus, Neumann, Tafjord, Dasigi, Liu,
  Peters, Schmitz, and Zettlemoyer}]{gardner-etal-2018-allennlp}
Matt Gardner, Joel Grus, Mark Neumann, Oyvind Tafjord, Pradeep Dasigi,
  Nelson~F. Liu, Matthew Peters, Michael Schmitz, and Luke Zettlemoyer. 2018.
\newblock \href {https://doi.org/10.18653/v1/W18-2501} {{A}llen{NLP}: A deep
  semantic natural language processing platform}.
\newblock In \emph{Proceedings of Workshop for {NLP} Open Source Software
  ({NLP}-{OSS})}, pages 1--6, Melbourne, Australia. Association for
  Computational Linguistics.

\bibitem[{Goodman(2020)}]{goodman2020penman}
Michael~Wayne Goodman. 2020.
\newblock Penman: An open-source library and tool for amr graphs.
\newblock In \emph{Proceedings of the 58th Annual Meeting of the Association
  for Computational Linguistics: System Demonstrations}, pages 312--319.

\bibitem[{He and Choi(2021)}]{he-choi:2021:iwpt}
Han He and Jinho~D. Choi. 2021.
\newblock Levi graph amr parser using heterogeneous attention.
\newblock In \emph{Proceedings of the 17th International Conference on Parsing
  Technologies and the IWPT 2021 Shared Task on Parsing into Enhanced Universal
  Dependencies}. Association for Computational Linguistics.

\bibitem[{He et~al.(2018)He, Lee, Levy, and Zettlemoyer}]{he-etal-2018-jointly}
Luheng He, Kenton Lee, Omer Levy, and Luke Zettlemoyer. 2018.
\newblock \href {https://doi.org/10.18653/v1/P18-2058} {{Jointly Predicting
  Predicates and Arguments in Neural Semantic Role Labeling}}.
\newblock In \emph{Proceedings of the 56th Annual Meeting of the Association
  for Computational Linguistics (Volume 2: Short Papers)}, pages 364--369,
  Melbourne, Australia. Association for Computational Linguistics.

\bibitem[{Jiang et~al.(2020)Jiang, Xu, Araki, and
  Neubig}]{jiang-etal-2020-generalizing}
Zhengbao Jiang, Wei Xu, Jun Araki, and Graham Neubig. 2020.
\newblock \href {https://doi.org/10.18653/v1/2020.acl-main.192} {Generalizing
  natural language analysis through span-relation representations}.
\newblock In \emph{Proceedings of the 58th Annual Meeting of the Association
  for Computational Linguistics}, pages 2120--2133, Online. Association for
  Computational Linguistics.

\bibitem[{Joshi et~al.(2020)Joshi, Chen, Liu, Weld, Zettlemoyer, and
  Levy}]{joshi-etal-2020-spanbert}
Mandar Joshi, Danqi Chen, Yinhan Liu, Daniel~S. Weld, Luke Zettlemoyer, and
  Omer Levy. 2020.
\newblock \href {https://doi.org/10.1162/tacl_a_00300} {{S}pan{BERT}: Improving
  pre-training by representing and predicting spans}.
\newblock \emph{Transactions of the Association for Computational Linguistics},
  8:64--77.

\bibitem[{Joshi et~al.(2019)Joshi, Levy, Zettlemoyer, and
  Weld}]{joshi-etal-2019-bert}
Mandar Joshi, Omer Levy, Luke Zettlemoyer, and Daniel Weld. 2019.
\newblock \href {https://doi.org/10.18653/v1/D19-1588} {{BERT} for coreference
  resolution: Baselines and analysis}.
\newblock In \emph{Proceedings of the 2019 Conference on Empirical Methods in
  Natural Language Processing and the 9th International Joint Conference on
  Natural Language Processing (EMNLP-IJCNLP)}, pages 5803--5808, Hong Kong,
  China. Association for Computational Linguistics.

\bibitem[{Judge et~al.(2006)Judge, Cahill, and
  Van~Genabith}]{judge2006questionbank}
John Judge, Aoife Cahill, and Josef Van~Genabith. 2006.
\newblock Questionbank: Creating a corpus of parse-annotated questions.
\newblock In \emph{Proceedings of the 21st International Conference on
  Computational Linguistics and 44th Annual Meeting of the Association for
  Computational Linguistics}, pages 497--504.

\bibitem[{Knight et~al.(2020)Knight, Badarau, Banarescu, Bonial, Bardocz,
  Griffitt, Hermjakob, Marcu, Palmer, O'Gorman et~al.}]{knight2020abstract}
Kevin Knight, Bianca Badarau, Laura Banarescu, Claire Bonial, Madalina Bardocz,
  Kira Griffitt, Ulf Hermjakob, Daniel Marcu, Martha Palmer, Tim O'Gorman,
  et~al. 2020.
\newblock Abstract meaning representation (amr) annotation release 3.0.
\newblock Technical report, Technical Report LDC2020T02, Linguistic Data
  Consortium, Philadelphia, PA, June.

\bibitem[{Kondratyuk and Straka(2019)}]{kondratyuk-straka-2019-75}
Dan Kondratyuk and Milan Straka. 2019.
\newblock \href {https://doi.org/10.18653/v1/D19-1279} {75 languages, 1 model:
  Parsing {U}niversal {D}ependencies universally}.
\newblock In \emph{Proceedings of the 2019 Conference on Empirical Methods in
  Natural Language Processing and the 9th International Joint Conference on
  Natural Language Processing (EMNLP-IJCNLP)}, pages 2779--2795, Hong Kong,
  China. Association for Computational Linguistics.

\bibitem[{Li et~al.(2019)Li, He, Zhao, Zhang, Zhang, Zhou, and
  Zhou}]{li2019dependency}
Zuchao Li, Shexia He, Hai Zhao, Yiqing Zhang, Zhuosheng Zhang, Xi~Zhou, and
  Xiang Zhou. 2019.
\newblock Dependency or span, end-to-end uniform semantic role labeling.
\newblock In \emph{Proceedings of the AAAI Conference on Artificial
  Intelligence}, volume~33, pages 6730--6737.

\bibitem[{Liu et~al.(2019)Liu, Ott, Goyal, Du, Joshi, Chen, Levy, Lewis,
  Zettlemoyer, and Stoyanov}]{liu2019roberta}
Yinhan Liu, Myle Ott, Naman Goyal, Jingfei Du, Mandar Joshi, Danqi Chen, Omer
  Levy, Mike Lewis, Luke Zettlemoyer, and Veselin Stoyanov. 2019.
\newblock Roberta: A robustly optimized bert pretraining approach.
\newblock \emph{arXiv preprint arXiv:1907.11692}.

\bibitem[{Manning et~al.(2014)Manning, Surdeanu, Bauer, Finkel, Bethard, and
  McClosky}]{manning2014stanford}
Christopher~D Manning, Mihai Surdeanu, John Bauer, Jenny~Rose Finkel, Steven
  Bethard, and David McClosky. 2014.
\newblock The stanford corenlp natural language processing toolkit.
\newblock In \emph{Proceedings of 52nd annual meeting of the association for
  computational linguistics: system demonstrations}, pages 55--60.

\bibitem[{M{\"u}ller et~al.(2015)M{\"u}ller, Cotterell, Fraser, and
  Sch{\"u}tze}]{muller2015joint}
Thomas M{\"u}ller, Ryan Cotterell, Alexander Fraser, and Hinrich Sch{\"u}tze.
  2015.
\newblock Joint lemmatization and morphological tagging with lemming.
\newblock In \emph{Proceedings of the 2015 Conference on Empirical Methods in
  Natural Language Processing}, pages 2268--2274.

\bibitem[{Qi et~al.(2020)Qi, Zhang, Zhang, Bolton, and
  Manning}]{qi-etal-2020-stanza}
Peng Qi, Yuhao Zhang, Yuhui Zhang, Jason Bolton, and Christopher~D. Manning.
  2020.
\newblock \href {https://doi.org/10.18653/v1/2020.acl-demos.14} {{S}tanza: A
  python natural language processing toolkit for many human languages}.
\newblock In \emph{Proceedings of the 58th Annual Meeting of the Association
  for Computational Linguistics: System Demonstrations}, pages 101--108,
  Online. Association for Computational Linguistics.

\bibitem[{Santorini(1990)}]{santorini1990part}
Beatrice Santorini. 1990.
\newblock Part-of-speech tagging guidelines for the penn treebank project (3rd
  revision).
\newblock \emph{Technical Reports (CIS)}, page 570.

\bibitem[{Song et~al.(2019)Song, Fore, Strassel, Lee, and
  Wright}]{song2019bolt}
Zhiyi Song, Dana Fore, Stephanie Strassel, Haejoong Lee, and Jonathan Wright.
  2019.
\newblock Bolt english sms/chat.

\bibitem[{Straka and Strakov{\'a}(2017)}]{straka2017tokenizing}
Milan Straka and Jana Strakov{\'a}. 2017.
\newblock Tokenizing, pos tagging, lemmatizing and parsing ud 2.0 with udpipe.
\newblock In \emph{Proceedings of the CoNLL 2017 Shared Task: Multilingual
  Parsing from Raw Text to Universal Dependencies}, pages 88--99.

\bibitem[{Tracey et~al.(2019)Tracey, Lee, and Strassel}]{tracey2019bolt}
Jennifer Tracey, Haejoong Lee, and Stephanie Strassel. 2019.
\newblock Bolt english discussion forums.

\bibitem[{Weischedel et~al.(2013)Weischedel, Palmer, Marcus, Hovy, Pradhan,
  Ramshaw, Xue, Taylor, Kaufman, Franchini et~al.}]{weischedel2013ontonotes}
Ralph Weischedel, Martha Palmer, Mitchell Marcus, Eduard Hovy, Sameer Pradhan,
  Lance Ramshaw, Nianwen Xue, Ann Taylor, Jeff Kaufman, Michelle Franchini,
  et~al. 2013.
\newblock {Ontonotes release 5.0 ldc2013t19}.
\newblock \emph{Linguistic Data Consortium, Philadelphia, PA}.

\bibitem[{Xu and Choi(2020)}]{xu-choi-2020-revealing}
Liyan Xu and Jinho~D. Choi. 2020.
\newblock \href {https://doi.org/10.18653/v1/2020.emnlp-main.686} {Revealing
  the myth of higher-order inference in coreference resolution}.
\newblock In \emph{Proceedings of the 2020 Conference on Empirical Methods in
  Natural Language Processing (EMNLP)}, pages 8527--8533, Online. Association
  for Computational Linguistics.

\bibitem[{Yan et~al.(2021)Yan, Gui, Dai, Guo, Zhang, and Qiu}]{yan2021unified}
Hang Yan, Tao Gui, Junqi Dai, Qipeng Guo, Zheng Zhang, and Xipeng Qiu. 2021.
\newblock A unified generative framework for various ner subtasks.
\newblock \emph{arXiv preprint arXiv:2106.01223}.

\bibitem[{Yu et~al.(2020)Yu, Bohnet, and Poesio}]{yu-etal-2020-named}
Juntao Yu, Bernd Bohnet, and Massimo Poesio. 2020.
\newblock \href {https://doi.org/10.18653/v1/2020.acl-main.577} {Named entity
  recognition as dependency parsing}.
\newblock In \emph{Proceedings of the 58th Annual Meeting of the Association
  for Computational Linguistics}, pages 6470--6476, Online. Association for
  Computational Linguistics.

\bibitem[{Zhang et~al.(2020)Zhang, Zhou, and Li}]{ijcai2020-560}
Yu~Zhang, Houquan Zhou, and Zhenghua Li. 2020.
\newblock \href {https://doi.org/10.24963/ijcai.2020/560} {{Fast and Accurate
  Neural CRF Constituency Parsing}}.
\newblock In \emph{Proceedings of the Twenty-Ninth International Joint
  Conference on Artificial Intelligence, {IJCAI-20}}, pages 4046--4053.
  International Joint Conferences on Artificial Intelligence Organization.
\newblock Main track.

\end{thebibliography}
\bibliographystyle{acl_natbib}

\end{document}